\documentclass[lettersize,journal]{IEEEtran}
\usepackage[numbers]{natbib}
\usepackage{caption}
\usepackage{setspace}
\usepackage{amssymb}
\usepackage{multirow}
\usepackage{lscape}
\usepackage{amsthm}
\usepackage{amsmath}
\usepackage{amssymb}
\usepackage{mathrsfs}
\usepackage{longtable}
\usepackage{lscape}
\usepackage{url}
\usepackage{tabularx}
\usepackage{makecell}
\usepackage{epsfig}
\usepackage{colortbl}
\usepackage{subfigure}
\usepackage[table]{xcolor}
\usepackage{amsmath,amssymb,amsfonts}
\DeclareCaptionLabelFormat{cont}{#1~#2\alph{ContinuedFloat}}
\usepackage{graphicx}
\usepackage{textcomp}
\usepackage{booktabs}  
\usepackage{threeparttable} 
\usepackage{array}
\usepackage{epstopdf}
\usepackage{epsfig}
\usepackage{graphicx}
\usepackage{stfloats}
\usepackage{graphicx}
\usepackage{textcomp}
\usepackage{diagbox}
\usepackage{float}
\usepackage{subfigure}
\usepackage{color}
\usepackage{amsthm} 
\usepackage{algorithm}
\usepackage{algorithmicx}
\usepackage{algpseudocode}

\newtheorem{definition}{Definition}

\begin{document}

\title{Multi-Domain Evolutionary Optimization on Combinatorial Problems in Complex Networks}

\author{Jie Zhao, Kang Hao Cheong,~\IEEEmembership{Senior Member, IEEE}, Yaochu Jin,~\IEEEmembership{Fellow, IEEE}
\thanks{
This work was supported by the Singapore Ministry of Education (MOE) Science of Learning, under Grant No. MOESOL2022-0003.

Jie Zhao and Kang Hao Cheong are affiliated with the Division of Mathematical Sciences, School of Physical and Mathematical Sciences, Nanyang Technological University, S637371, Singapore. Kang Hao Cheong is also with the College of Computing and Data Science, Nanyang Technological University, S639798, Singapore.

Yaochu Jin is affiliated with the School of Engineering, Westlake University, Hangzhou, 310030, China.

Corresponding Author: K.H. Cheong (kanghao.cheong@ntu.edu.sg).}
}

\markboth{Journal of \LaTeX\ Class Files,~Vol.~14, No.~8, August~2021}%
{Shell \MakeLowercase{\textit{\textit{et al.}}}: A Sample Article Using IEEEtran.cls for IEEE Journals}


\maketitle

\begin{abstract}
Knowledge transfer-based evolutionary optimization has garnered significant attention, such as in multi-task evolutionary optimization (MTEO), which aims to solve complex problems by simultaneously optimizing multiple tasks. While this emerging paradigm has been primarily focusing on task similarity, there remains a hugely untapped potential in harnessing the shared characteristics between different domains. For example, real-world complex systems usually share the same characteristics, such as the power-law rule, small-world property and community structure, thus making it possible to transfer solutions optimized in one system to another to facilitate the optimization. Drawing inspiration from this observation of shared characteristics within complex systems, we present a novel framework, multi-domain evolutionary optimization (MDEO). First, we propose a community-level measurement of graph similarity to manage the knowledge transfer among domains. Furthermore, we develop a graph learning-based network alignment model that serves as the conduit for effectively transferring solutions between different domains. Moreover, we devise a self-adaptive mechanism to determine the number of transferred solutions from different domains, and introduce a knowledge-guided mutation mechanism that adaptively redefines mutation candidates to facilitate the utilization of knowledge from other domains. To evaluate its performance, we use a challenging combinatorial problem known as adversarial link perturbation as the primary illustrative optimization task. Experiments on multiple real-world networks of different domains demonstrate the superiority of the proposed framework in efficacy compared to classical evolutionary optimization. 

\end{abstract}


\section{Introduction}

\IEEEPARstart{K}{nowledge} transfer in evolutionary optimization has emerged as a promising paradigm for tackling complex optimization problems by enabling the sharing of useful information across related problems. One representative framework is multi-task evolutionary optimization (MTEO), which has seen notable progress in recent years \cite{chen2022scaling,al2023multitree}, following the introduction of the multi-factorial evolutionary algorithm (MFEA) \cite{gupta2015multifactorial}. MTEO is designed to handle scenarios involving multiple related tasks being optimized concurrently, with the aim of leveraging inter-task similarity to improve performance on each task \cite{shakeri2022scalable,bai2021multitask}.

\subsection{{Motivation}}

In this work, we shift the focus from task-based settings to the structural properties of the network: While most existing endeavors primarily concentrate on exploring relationships between tasks, a research gap persists regarding correlations between different domains in evolutionary optimization. In real-world applications, we encounter a multitude of networks representing different complex systems \cite{zhao2023early}. These networks may represent social networks \cite{wen2024eriue} or transportation networks \cite{wen2022exploring}, many of which share common structural properties, such as power-law distribution \cite{clauset2009power,zhao2021complex}, community structure \cite{wen2021gravity,zhao2024swarm} and small-world characteristic \cite{barabasi1999emergence,watts1998collective,zhao2024mase}. In real-world scenarios, the same objective often exists in different domains, such as the task of critical node mining in complex networks \cite{wen2024eriue,wen2020identification,zhao2025visual}, which can contribute to the prevention of catastrophic outages in power networks \cite{motter2002cascade}, the identification of drug target candidates in protein networks \cite{csermely2013structure}, or the improvement of the robustness in communication networks \cite{albert2000error}. The existence of shared attributes and structural properties across real-world networks presents an avenue for knowledge transfer and collaborative optimization when dealing with an identical task across different systems. 

To date, MTEO in complex networks has already been extensively studied \cite{wu2021evolutionary,wang2023multi}, however, leveraging the correlation of domains to facilitate the optimization of network structure remains underexplored. Therefore, we propose a new framework called multi-domain evolutionary optimization (MDEO) for network-structured combinatorial problems, in which solutions obtained from optimizing one network can be effectively transferred and adapted to optimize another network with improved efficacy. The following are the main differences between MTEO and MDEO, justifying the novelty of our work:

\textbf{Granularity of Tasks}: {In MTEO, the emphasis is on optimizing different tasks within a single domain. Conversely, MDEO focuses on optimizing an identical task in multiple networks representing different complex systems. The goal of MDEO is to find the optimal solution for each domain. Note that if multiple tasks concurrently exist across different domains, MDEO could be extended to multi-domain multi-task evolutionary optimization.}

\textbf{Knowledge Transfer Across different domains}: {In MDEO, knowledge is transferred across distinct domains, such as the social network, power network and biology network, enhancing solutions by leveraging insights from one domain to another. This contrasts with MTEO, where knowledge is transferred within the same domain across tasks with similar characteristics, whether continuous or combinatorial.}

\subsection{{Contributions}}

In this study, we take a further stride and delve into the case of more than three networks, i.e., many networks will be involved in the optimization. The proposed MDEO consists of four components 1) graph similarity; 2) graph embedding; 3) network alignment; 4) many-network evolutionary optimization and its overview is as follows:

Multi-domain evolutionary optimization may present a significant challenge due to the inherent computational complexity when dealing with a large number of networks. Additionally, the efficacy of solutions on one network may not necessarily translate to success on others. To address these issues, we propose a new measurement of \textbf{graph similarity}, quantifying the closeness between networks at the community level. The similarity measurement enables us to focus on a subset of closely related networks during the knowledge transfer process, thereby effectively reducing the computational burden and increasing the likelihood of successful adaptations.

In the multi-domain context, transferring solutions across different networks necessitates the establishment of node correspondences. We first employ the graph autoencoder to obtain \textbf{graph embeddings} aimed at capturing node similarity and higher-order interactions. Building upon the derived graph embeddings, we then propose a novel \textbf{network alignment} method that combines supervised and unsupervised learning. In supervised learning, we propose a community-level anchor node selection method to build up the training set and improve the alignment accuracy. This approach empowers us to create mappings of nodes sharing analogous roles or structures across different networks.

In the process of \textbf{many-network evolutionary optimization}, we transfer solutions from similar networks to the target network, achieved by the node mappings obtained through network alignment. By observing the contributions of these transferred solutions, we leverage a self-adaptive model to determine the appropriate number of solutions to transfer between networks, with the consideration of the previously calculated graph similarity. 

Validation on eight real-world networks for edge-level tasks reveals that the MDEO framework achieves higher average fitness than classical evolutionary optimization. {This trend is also observed in the node-level tasks, indicating the generalizability and adaptability of MDEO.}


\subsection{{Organization}}

The subsequent sections of the paper are organized as follows: Section \ref{Problem definition} provides an overview of related work, and Section \ref{formulation} introduces the optimization problem--community deception. Section \ref{method} delves into the intricacies of our proposed MDEO framework, detailing its methodology and components. Section \ref{experiment} entails an examination of the effectiveness of our method through a series of experiments conducted on various real-world networks. {Section \ref{sec.discussion} discusses the possible extension of MDEO.} Lastly, Section \ref{conclusion} concludes with a summary of our main results.

\section{Related work}\label{Problem definition}

The capability of evolutionary optimization in handling discrete problems with non-linear characteristics has led to their extensive utilization within complex systems. For instance, Wang \textit{et al.} \cite{wang2021computationally} developed a multi-objective model by considering the nonuniform latency and computational complexity to enhance the tolerance of networks against attacks. Similarly, Wu \textit{et al.} \cite{wu2020evolutionary} leveraged the community distributions of networks to downsize the search space and introduce a multi-objective framework for network reconstruction by optimizing the reconstruction error and sparsity. Based on the same optimization objectives, Ying \textit{et al.} \cite{ying2021multiobjective} proposed a logistic principal component analysis-based method to improve the representation of networks. {Moreover, evolutionary optimization has also been used in the diffusion source localization for sensor deployment to enhance the identification accuracy \cite{zhao2025enhanced}.}

Meanwhile, evolutionary algorithms were utilized to address the problem of community deception as well. Chen \textit{et al.} \cite{chen2019ga} used the genetic algorithm to minimize the modularity and suggest the modifications to edges. Further, they categorized this problem into node communities, target communities, and all communities according to the scales \cite{chen2020multiscale}. Subsequently, a self-adaptive evolutionary framework is designed to streamline the search for the optimal edge set, complemented by the development of a permanence-based method to minimize the search space \cite{zhao2023self}. To mitigate the costs associated with evaluation, Zhao and Cheong \cite{zhao2023Obfuscating} developed a divide-and-conquer strategy that partitions the network into sub-components, each optimized individually and cooperatively using the evolutionary algorithm. More applications of evolutionary algorithms on complex systems include dynamic community detection \cite{ma2023higher}, and module identification \cite{chen2019mumi}. These applications demonstrate the versatility and effectiveness of evolutionary algorithms in tackling diverse challenges within the field of complex network analysis. 

Recently, considerable efforts have been directed towards tackling evolutionary multi-task cases where  the development of multi-factorial evolutionary algorithm \cite{gupta2015multifactorial} has laid the foundation for related studies. In addition, the optimization on single objective has been extended to the case of multiple objectives \cite{gupta2016multiobjective}, adopting the implicit transfer mode where the crossover is operated on the solutions aiming at different tasks \cite{tang2018group,lin2020effective}. \textcolor{black}{In another extension of MTEO to the multi-objective setting \cite{lin2023multiobjective}, each task is decomposed into multiple subproblems, and the transfer probability of a solution is assessed based on how much it improves the performance of its corresponding subproblem.} There is another route referred to explicit transfer that uses task-specific information to guide knowledge exchange between tasks by using statistical information \cite{lin2019multiobjective} or constructing a mapping matrix \cite{feng2020explicit,feng2018evolutionary}. \textcolor{black}{The primary focus of MTEO research lies in enhancing transfer strategies. For example, Wu \textit{et al.} \cite{wu2024diversified} proposed a diversified reasoning approach that expands solution space via varied transfer patterns. Xue \textit{et al.} \cite{xue2024neural} used neural networks to evaluate task similarity and guide information flow, thereby enhancing transfer quality. In \cite{wu2025learning}, reinforcement learning is used to learn effective policies where the knowledge transfer is modeled as sequential decisions.}

\begin{table*}[ht]
\centering
\caption{{Comparison of evolutionary optimization frameworks based on knowledge transfer. Multi-task evolutionary optimization is divided into two categories: MTEO-ConO, which addresses continuous problems, and MTEO-ComO, which focuses on combinatorial problems.}}
\label{tab:comparison}
\begin{tabular}{@{}lcccc@{}}
\Xhline{5\arrayrulewidth}
\textbf{Framework} & \textbf{Domain Scope} & \textbf{Problem Type} & \textbf{Space Type} & \textbf{Task Scope} \\ \midrule
{\textbf{MTEO-ConO} \cite{gupta2015multifactorial,gupta2016multiobjective,tang2018group,lin2020effective,qiao2023self,jiang2023block}} & {Single-domain} & {Continuous optimization} & {Continuous} & {Multiple} \\
{\textcolor{black}{\textbf{MTEO-ComO} }\cite{wu2021evolutionary,feng2020explicit,zhou2016evolutionary,huang2023evolutionary,wang2023enhancing,wang2023multia,yang2025evolutionary}} & {Single-domain} & {Combinatorial optimization} & {Discrete} & {Multiple} \\
\textbf{MDEO} & {Multi-domain} & {Combinatorial optimization} & {Discrete} & {Single/Multiple} \\ \Xhline{5\arrayrulewidth}
\end{tabular}
\end{table*}

Beyond conventional areas, MTEO has been leveraged to strengthen the robustness of complex networks. For example, Wang \textit{et al.} \cite{wang2023multi} utilized the correlation of tasks to develop a new crossover operator, thereby injecting knowledge into different individuals. According to \cite{wang2023enhancing}, a graph neural network-based method has been developed that facilitates the exchange of information, demonstrating that structural destruction and cascading failure, though seemingly unrelated goals, can mutually enhance optimization outcomes. In a study on influence maximization \cite{wang2023multia}, each task is formulated as a transformation, allowing for the effective application of MTEO and resulting in enhanced performance. Wu \textit{et al.} \cite{wu2021evolutionary} promoted the reconstruction of multiplex networks using MTEO with the inherent correlation of different layers. Similar research by Lyu \textit{et al.} \cite{lyu2022community} formulates the modularity optimization within each layer as an independent task, facilitating community detection in multiplex networks.

{Even though networks typically display common patterns, there has been minimal focus on cross-domain evolutionary optimization, especially with respect to network structures. Motivated by this overlooked potential, we introduce a novel optimization framework---multi-domain evolutionary optimization (MDEO). The comparison of the proposed MDEO and the existing MTEO literature is shown in Table \ref{tab:comparison}.}

\section{Optimization problem formulation}\label{formulation}
As a security-related task \cite{waniek2018hiding,fionda2017community,hao2022iron}, community deception involves nodes intentionally concealing their true community affiliations by modifying (adding and removing) connections, which serves as the illustration of our MDEO for the following concerns: 

\textbf{Exploiting Community Detection Algorithms:} With the assistance of community detection algorithms, densely interconnected groups of nodes with high intra-group connectivity and low inter-group connectivity can be easily identified \cite{lyu2022community,xia2021fast}. The development of community detection, while useful for understanding network structures and interactions, can also have certain dangerous aspects caused by the potential misuse or exploitation of community information \cite{wondracek2010practical,remy2018tracking}.

\textbf{Complexity of Community Deception:} As an edge-level task, the search space of community deception is $|V|^2$, much higher than node-level tasks of $|V|$. Both existing and nonexistent edges will be considered in community deception, meaning the genes (edges) in the chromosomes (edge set) are heterogeneous. Therefore, community deception presents additional challenges than traditional tasks in complex networks such as the aforementioned critical node mining.

Community deception serves as an innovative advancement in network science, radically altering node affiliations through minimal topological adjustments. Specifically, the modification of the network involves the addition of nonexistent edges and the deletion of existing edges. Suppose $G = \{V, E\}$ is a network where $V$ and $E$ denote the nodes and edges, respectively. The modified edges can be defined as follows:

\begin{equation}
E'=\left(E \cup E^{+}\right) \backslash E^{-},
\end{equation}
where $E^{+}$ and $E^{-}$ denote the edges to add and delete, i.e.,

\begin{equation}
\begin{array}{l}
E^{+} \subseteq\{(u, v): u \in V \wedge v \in V,(u, v) \notin E\}, \\
E^{-} \subseteq\{(u, v): u \in V \wedge v \in V,(u, v) \in E\},
\end{array}
\end{equation}
with
\begin{equation}
\left|E^{+}\right|+\left|E^{-}\right| \leq \beta,
\end{equation}
where $\beta$ is the perturbance budget, referring to the maximum number of edges to rewire.

Assume there are two community structures belonging to the original network $G$ and modified network $G'$, $\widetilde{\mathcal{C}}= \{\mathcal{C}_1, \mathcal{C}_2,\cdots,\mathcal{C}_k \}$ and $\widetilde{\mathcal{C}'}= \{\mathcal{C}'_1, \mathcal{C}'_2,\cdots,\mathcal{C}'_{k'} \}$, respectively, the problem of community deception is defined as:

\begin{equation}
\underset{\left\{E^+, E^-\right\}}{\arg \max }\hspace{0.5em}\{\phi(\widetilde{\mathcal{C}}, \widetilde{\mathcal{C}')},E^+, E^-\}.
\end{equation}

The function $\phi$ refers to the disparity between the community structures before and after modification. To validate the generalizability of MDEO, the node-level task is also tested in  Section \ref{Sec.ge}.

\section{Multi-Domain evolutionary optimization}\label{method}
This section provides a detailed introduction to the proposed MDEO, with its diagram shown in Figure \ref{flowchart}. {The mathematical notations used in MDEO are listed in Table \ref{notation}.}

\begin{table}[ht]
\centering
\caption{{Mathematical notations.}}
\resizebox{88mm}{28mm}{
\begin{tabular}{cc}
\Xhline{5\arrayrulewidth}
 \textbf{Symbols} & \textbf{Definition} \\
\hline
{$G$} & {A network consisting of nodes $V$ and edges $E$} \\
{$\mathcal{G}$} & {Set of networks for optimization} \\
{$\widetilde{\mathcal{C}}$} & {Community distribution of the network $G$} \\
{$\beta$} & {Modification budget for the network $G$} \\
{$\mathcal{A}$} & {Set of anchor nodes in the networks} \\
{$v^{X,l}_i$} & {The $i$-th largest-degree node in the node set $X$} \\
{$v^{X,s}_i$} & {The $i$-th smallest-degree node in the node set $X$} \\
{$\Phi^{i\rightarrow j}$} & {Embedding mapping from network $G_i$ to network $G_j$} \\
{$M_{i \rightarrow j}$} & {Edge mapping from network $G_i$ to network $G_j$} \\
{$\mathcal{A}^{\widetilde{\mathcal{C}}}_{A,B}$} & {Aligned community between networks $G_i$ and $G_j$} \\
{$\mathcal{S}^\mathcal{G}_{i,j}$} & {Similarity bwtween networks $G_i$ and $G_j$} \\
{$T_{i \leftarrow j}$} & {Solution transferred from network $G_j$ to network $G_i$} \\
\Xhline{5\arrayrulewidth}
\end{tabular}}
\label{notation}
\end{table}

\begin{figure*}[htbp]
\centering
\includegraphics[height=10cm,width=17cm]{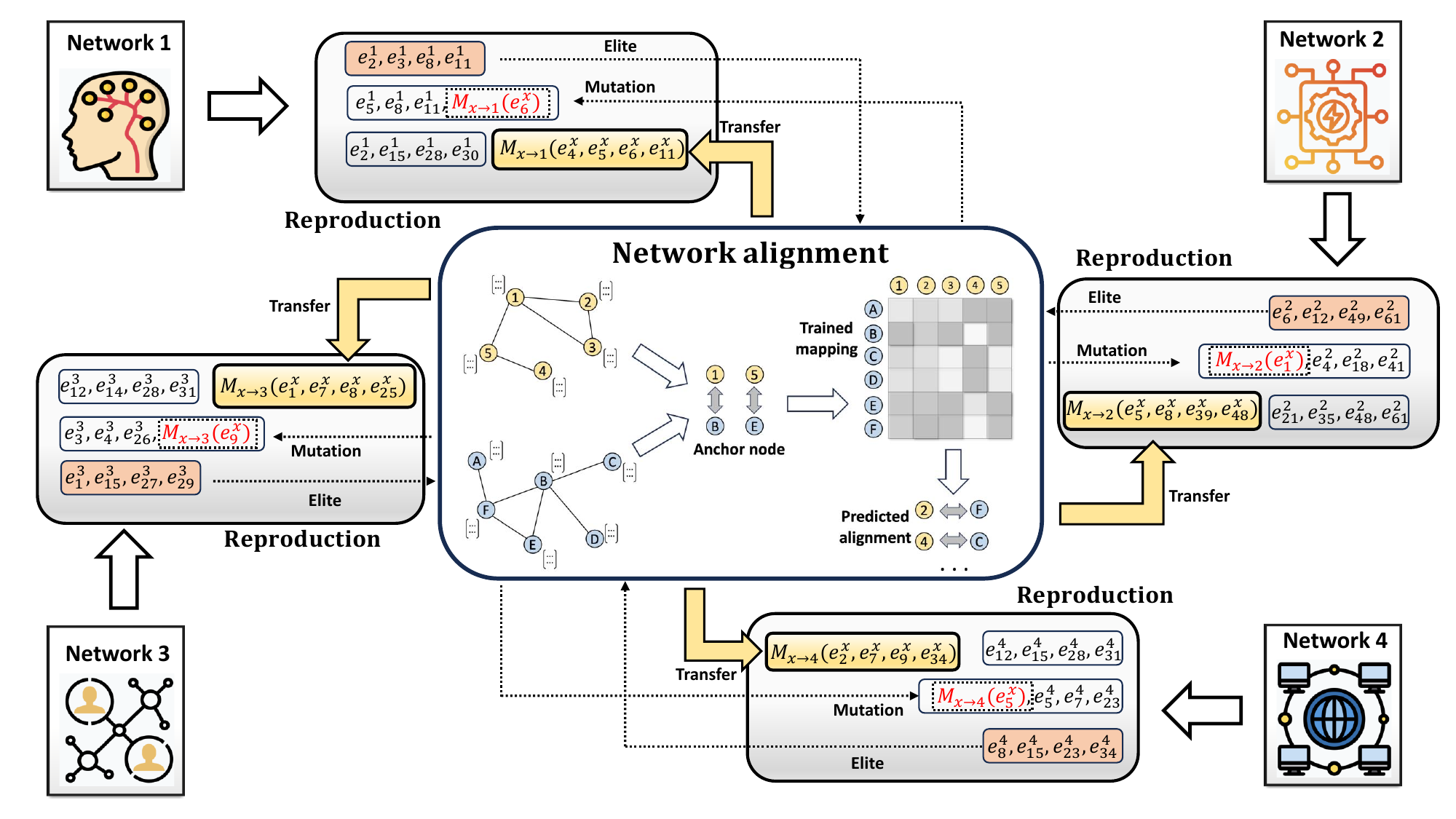}
\caption{The diagram of multi-domain evolutionary optimization. Four networks representing different systems (brain network, power network, social network, communication network) are optimized simultaneously. The transfer of knowledge across networks is achieved through the trained network alignment model. $e^{i}_j$ denote the $j$-th edge in $i$-th network and $M_{i \rightarrow j}$ is the edge mapping from the $i$-th network to the $j$-th network. {The solution shown in yellow represents the transferred solution, while the light red color represents the elite solution from other networks.} The red edge is the mutated edge that will be replaced by the edge of elite solutions of other networks. The knowledge exchange occurs when a series of conditions are met, and only elite solutions will be utilized to assist the optimization of other networks.} 
\label{flowchart}
\end{figure*}

\subsection{Graph similarity}
The measurement of graph similarity enables us to perform selective knowledge transfer between networks. Instead of transferring all available knowledge, it allows us to focus exclusively on pertinent domains, minimizing the risk of negative transfer and reducing computational complexity.

In general, the graph similarity can be easily obtained by averaging node similarity of networks \cite{wen2019node} or computing the distance of graph embeddings \cite{Ying2018hierarchical,gao2019graph}. However, the former approach oversimplifies the complex interactions within the graph, ignoring the community structure and global patterns of networks. On the other hand, existing graph pooling techniques are typically trained on small graphs, potentially constraining their efficacy on large-scale graphs.

Here, we propose a community-level measurement of graph similarity. Given a network $G = (V, E)$ and its community distribution $\widetilde{\mathcal{C}} = \left\{\mathcal{C}_{1}, \mathcal{C}_{2}, \ldots, \mathcal{C}_{k}\right\}$, for each community $\mathcal{C} \in \widetilde{\mathcal{C}}$, we can define $D_\mathcal{C} = \{d_v|v \in \mathcal{C}\}$ as the degree distribution of community $\mathcal{C}$ where $d_v$ is the node degree of $v$. With the consideration of the scales of different networks, we normalize this distribution by dividing the largest degree of the network, denoted as $D'_\mathcal{C} = \{d_v / d^{max}|v \in \mathcal{C}\}$ where $d^{max}$ is the largest degree in the network $G$. In addition, another challenge emerges due to the varying lengths of these normalized distributions across networks, which complicates the direct comparison of community degree distributions. To address this issue, we unify the degree distributions across communities of different scales into the same length.

\begin{definition}
\textbf{Degree interval-based distribution}: Given a community $\mathcal{C} \in \widetilde{\mathcal{C}}$, then we can obtain the normalized degree distribution $D'_\mathcal{C} = \left\{d'_1, d'_2, \ldots, d'_n\right\}$  where $n$ is the number of nodes in community. Let $I_k=\left[a_k, b_k\right)$ represent the $k$-th interval, where $b_k$ is the upper bound of the interval and $a_k$ is the lower bound, then we can have

\begin{equation}
I_j=[(b_j-a_j) * j, (b_j-a_j) * (j+1)).
\end{equation}

For $x_i \in D'_\mathcal{C}$, corresponding interval $I_j$ can be found such that:
$x_i \in I_j$ where $j=\left\lfloor\frac{x_i}{b_j-a_j}\right\rfloor$. Then, we can calculate the number of data points $x_i \in D'_\mathcal{C}$ that fall into each interval $I_k:$
$h_j=\operatorname{Count}\left(x_i\right.$ such that $\left.x_i \in I_j\right)$ for each $j$. Here, we set $b_j - a_j = 0.2$ for $j \in \{0,1,2,3,4\}$. The degree interval-based distribution can be obtained as follows:

\begin{equation}
H_\mathcal{C} = \{\frac{h_0}{|\mathcal{C}|}, \frac{h_1}{|\mathcal{C}|}, \cdots, \frac{h_l}{|\mathcal{C}|}\},
\end{equation}
where $l = 1/(b_j - a_j) - 1$. 
\end{definition}
With the above definition, we achieve a uniform representation of the degree distribution for each community, consistently comprising $l$ elements regardless of the number of nodes in the community.

Given two community structures $\widetilde{\mathcal{C}_A} = \{\mathcal{C}_1, \mathcal{C}_2\,\cdots,\mathcal{C}_k \}$ and $\widetilde{\mathcal{C}_B} = \{\mathcal{C}'_1, \mathcal{C}'_2\,\cdots,\mathcal{C}'_{k'} \}$, the similarity of communities $\mathcal{C} \in \widetilde{\mathcal{C}_A}, \mathcal{C}' \in \widetilde{\mathcal{C}_B}$ can be defined as follows:

\begin{align}
    \operatorname{Diff}\left(\mathcal{C}, \mathcal{C}'\right) &=  \frac{1}{2}\left(\mathbb{KL}\left(H_{\mathcal{C}} \| H_{\mathcal{C}'}\right)+\mathbb{KL}\left(H_{\mathcal{C}'} \| H_{\mathcal{C}}\right)\right) \nonumber \\
      &\quad \times \mathbf{Max}(|\mathcal{C}|/|\mathcal{C}'|, |\mathcal{C}'|/|\mathcal{C}|),
\end{align}
where $\mathbb{KL}(.)$ refers to the Kullback–Leibler divergence and the second item $\mathbf{Max}(|\mathcal{C}|/|\mathcal{C}'|, |\mathcal{C}'|/|\mathcal{C}|)$ is used to quantify the disparity in community sizes.

Traditional graph representations focus on pairwise relationships between nodes while higher-order graphs extend this concept and can capture interactions involving more than two nodes. Therefore, we also take into account higher-order network knowledge, enabling a more refined analysis of structural information. Given a network $G$ and a motif $M$, the network $G$ can be converted into a weighted network $G_M$ as follows:

\begin{equation}
{G}^{{M}}=\left\{{V}, {E}^{{M}}\right\},
\end{equation}
where ${E}^{M} \in E$ represents the set of edges of ${G}^{M}$.  We can measure the higher-order difference between $\mathcal{C}$ and $\mathcal{C}'$ as follows:

\begin{align}
    \operatorname{Diff}^M\left(\mathcal{C}, \mathcal{C}'\right) &=  \frac{1}{2}\left(\mathbb{KL}\left(H^M_{\mathcal{C}} \| H^M_{\mathcal{C}'}\right)+\mathbb{KL}\left(H^M_{\mathcal{C}'} \| H^M_{\mathcal{C}}\right)\right) \nonumber \\
      &\quad \times \mathbf{Max}(|\mathcal{C}|/|\mathcal{C}'|, |\mathcal{C}'|/|\mathcal{C}|).
\end{align}

The similarity between communities is defined as follows:

\begin{equation}
\mathcal{S}(\mathcal{C},\mathcal{C}') = \textbf{Exp} (-\operatorname{Diff}\left(\mathcal{C}, \mathcal{C}'\right)) + \textbf{Exp} (-\operatorname{Diff}^M\left(\mathcal{C}, \mathcal{C}'\right)).
\end{equation}

\begin{definition}
\textbf{Aligned community}: Given two networks $G_A$, $G_B$ and their corresponding community structures $\widetilde{\mathcal{C}_A} = \{\mathcal{C}_1, \mathcal{C}_2,\cdots,\mathcal{C}_k \}$ and $\widetilde{\mathcal{C}_B} = \{\mathcal{C}'_1, \mathcal{C}'_2,\cdots,\mathcal{C}'_{k'} \}$, we can have the community-wise similarity matrix $\mathcal{S}_{i,j}  \in \mathbb{R}^{k \times k'}$ where $ i \in \{1,2,\cdots,k\}, j \in \{1,2,\cdots,k'\}$. Then, we define the set of aligned communities as follows:
\begin{equation}
\mathcal{A}^{\widetilde{\mathcal{C}}}_{A,B} = \{\{\mathcal{C}_{a_1}, \mathcal{C}'_{b_1}\},\{\mathcal{C}_{a_2}, \mathcal{C}'_{b_2}\}, \cdots, \{\mathcal{C}_{a_z}, \mathcal{C}'_{b_z}\}\},
\end{equation}
where $z = \mathbf{Min}(|\widetilde{\mathcal{C}_A}|,|\widetilde{\mathcal{C}_B}|)$. $\mathcal{C}_{a_i}$ and $\mathcal{C}'_{b_i}$ refer to the $a_i$-th and $b_i$-th community in $\widetilde{\mathcal{C}_A}$ and $\widetilde{\mathcal{C}_B}$, respectively.
Accordingly, we can have the set of their corresponding similarity values as follows:
\begin{equation}
\mathcal{S}^{\widetilde{\mathcal{C}}}_{A,B} = \{\mathcal{S}_{a_1,b_1},\mathcal{S}_{a_2,b_2}, \cdots, \mathcal{S}_{a_z,b_z}\}, 
\end{equation}
which satisfies

\begin{equation}
\mathcal{S}_{a_i,b_i} = \mathbf{Max}(\mathcal{S}/ \bigcup_{i > j} (\mathcal{S}_{a_j,} \cup \mathcal{S}_{,b_j})).
\end{equation}
\end{definition}

The similarity between $G_A$ and $G_B$ is defined as the average similarity of aligned community pairs.
\begin{equation}
\mathcal{S}^\mathcal{G}_{A,B} = \frac{1}{\mathbf{Min}(|\widetilde{\mathcal{C}_A}|,|\widetilde{\mathcal{C}_B}|)} \sum_{x \in \mathcal{S}^{\widetilde{\mathcal{C}}}_{A,B}} x.
\label{graph_similarity}
\end{equation}

{The graph similarity matrix \(\mathcal{S}^\mathcal{G} \) is initialized based on Eq. \ref{graph_similarity}. Initially, it is used to identify the assisted network for each network and is subsequently updated to manage the transfer of solutions from each assisted network.
} The process of calculating the graph similarity is presented in Algorithm \ref{a2}.

\begin{algorithm}[t]
\caption{Graph Similarity} 
\hspace*{0.02in} {\bf Input:} 
Networks $G_A = (V_A, E_A)$ and $G_B = (V_B, E_B)$\\
\hspace*{0.02in} {\bf Output:} 
The graph similarity between $G_A$ and $G_B$
\begin{algorithmic}[1]
\State  Obtain higher-order graphs ${G}^{M}_A= ({V}_A, {E}^{M}_A)$ and ${G}^{M}_B=(V_B, {E}^{{M}}_B)$
\State  Identify the community of $G_A$ and $G_B$ to obtain $\widetilde{\mathcal{C}}_A = \{\mathcal{C}_1, \mathcal{C}_2\,\cdots,\mathcal{C}_k \}$ and $\widetilde{\mathcal{C}}_B = \{\mathcal{C}'_1, \mathcal{C}'_2\,\cdots,\mathcal{C}'_{k'} \}$
\State Compute the degree distribution of each community of $G$ and $G^{m}$ 
\State Normalize the degree distribution based on the scale of the network and transform it into the probability distribution
\State Construct the similarity matrix between the communities of the two networks based on K-L divergence
\For{$i = 1$ to $\mathbf{Min}\{k, k'\}$}
\State Find the largest value of $\mathcal{S}$ and its two corresponding communities as well as its index $i_{\text{max}}, j_{\text{max}} = \text{argmax} S_{i,j} $ where $ 1 \leq i \leq k, 1 \leq j \leq k'$
\State Update $\mathcal{S}^\mathcal{G}_{A,B} \leftarrow \mathcal{S}^\mathcal{G}_{A,B} + \mathbf{Max}(\mathcal{S})$
\State Set the $i_\text{max}$ row and $j_\text{max}$ column of $S$ to $0$
\EndFor
\State Normalize the similarity $\mathcal{S}^\mathcal{G}_{A,B}$ 
\end{algorithmic}
\label{a2}
\end{algorithm}

\subsection{Graph embedding}
To construct the node mappings across different networks, we employ graph autoencoders (GAE) to generate embeddings capturing meaningful and compact graph representations without the labeled data \cite{kipf2016variational}.
In GAE, the graph convolutional network (GCN) is taken as the encoder \cite{kipf2017semi}, i.e.,
\begin{equation}
\mathbf{Z}=f_{encoder}(G) = \operatorname{GCN}(\mathbf{X}, \mathbf{A}),
\end{equation}
where $\mathbf{A}$ denotes the adjacency matrix and $\mathbf{X} \in \mathbb{R}^{N \times d}$ denotes nodes features. The information aggregation of GCN is

\begin{equation}
\operatorname{GCN}(\mathbf{X}, \mathbf{A})=\widetilde{\mathbf{A}} \operatorname{ReLU}\left(\widetilde{\mathbf{A}} \mathbf{X} \mathbf{W}_{\mathbf{0}}\right) \mathbf{W}_{\mathbf{1}},
\end{equation}
where $\mathbf{W}_{\mathbf{0}}$ and $\mathbf{W}_{\mathbf{1}}$ are learnable parameters and $\widetilde{\mathbf{A}}=\mathbf{D}^{-\frac{1}{2}} \mathbf{A} \mathbf{D}^{-\frac{1}{2}}$.

The decoder takes the latent space representation $\mathbf{Z}$ as input to reconstruct the adjacency matrix. The inner product is adopted as the decoder function:

\begin{equation}
\hat{\mathbf{A}}=f_{decoder}(\mathbf{Z}) = \sigma\left(\mathbf{Z} \mathbf{Z}^{\mathrm{T}}\right),
\end{equation}
where $\sigma(\cdot)$ is the logistic sigmoid function.

GAE is trained by minimizing the reconstruction loss that measures the dissimilarity between the adjacency matrix $\mathbf{A}$ and the reconstructed matrix $\hat{\mathbf{A}}$. The loss function is:

\begin{equation}
\mathcal{L}_{reconstruct}=-\frac{1}{N} \sum y \log \hat{y}+(1-y) \log (1-\hat{y}),
\end{equation}
where $y$ refers to the element in adjacency matrix $\mathbf{A}$ and $\hat{y}$ is the element that has the same position with $y$ in $\hat{\mathbf{A}}$.


\subsection{Network alignment}
To achieve knowledge transfer across different networks, we propose a new dual-learning network alignment model. It aims to find a correspondence (mapping) between nodes that have similar roles across different networks. The pseudocode of training network alignment is shown in Algorithm \ref{a4}.

\begin{definition}
\textbf{Network alignment}: Given two networks $G_A = (V_A, E_A)$ and $G_B = (V_B, E_B)$, and their embeddings $\mathbb{E}_A$ and $\mathbb{E}_B$, network alignment model $\Phi$ is defined as a mapping from $G_A$ to $G_B$, such that for each $u \in V_A$ and its embedding $\mathbb{E}_A\left(u\right)$, we can have $\Phi^{a\rightarrow b}\left(\mathbb{E}_A\left(u\right)\right)=$ $\mathbb{E}_B\left(v\right), v \in V_B$  where $\{u,v\}$ is a pair of anchor nodes that play a similar role in their respective networks. We also denote the inverse mapping in a similar manner as $\Phi^{b\rightarrow a}\left(\mathbb{E}_B\left(v\right)\right)=\mathbb{E}_A\left(u\right)$.
\end{definition}

A collection of anchor nodes is carefully chosen based on their structural functions in both networks, serving as labels to train the network alignment model. Given two networks $G_A = (V_A, E_A)$ and $G_B = (V_B, E_B)$, we can then obtain their aligned community pairs $\mathcal{A}^{\widetilde{\mathcal{C}}}_{A,B}$ as in \textbf{Definition 2}, which is the basis of the construction of anchor nodes. To adapt the network alignment model for multi-domain scenarios, both trivial and important nodes within each aligned community pair will be considered as anchor nodes.

\begin{definition}
\textbf{Anchor node (large degree) -- $\mathcal{A}^{large}$}: Suppose the $i$-th largest-degree node in the set $X$ is $v^{X,l}_i$. Then, given two networks $G_A$ and $G_B$, and an aligned community pair $\{\mathcal{C}, \mathcal{C}'\} \in \mathcal{A}^{\widetilde{\mathcal{C}}}_{A,B}$, the set of anchor large-degree nodes can be formulated as follows:

\begin{equation}
\mathcal{A}^{large} = \bigcup\limits^{\{\mathcal{C}, \mathcal{C}'\} \in \mathcal{A}^{\widetilde{\mathcal{C}}}_{A,B}} \{v^{\mathcal{C},l}_i, v^{\mathcal{C}',l}_i\}, i \in \{1,...,k_l\},
\end{equation}
where $k_l = \mathbf{Min}\{log_2|\mathcal{C}|,log_2|\mathcal{C}'|\}$.

\end{definition}

The loss for aligning the important nodes can be obtained as follows:

\begin{equation}
\begin{split}
\mathcal{L}_{large} &= \sum\limits_{\{u,v\} \in \mathcal{A}^{large}} dist(\Phi^{a\rightarrow b}(\mathbb{E}_A(u)) , \mathbb{E}_B(v)) \\
&\quad + dist(\Phi^{b\rightarrow a}(\mathbb{E}_B(v)), \mathbb{E}_A(u)),
\end{split}
\end{equation}
where $dist$ refers to the mean square error (MSE), measuring the difference between two embeddings.

To preserve the structural information, we will not consider the nodes with the least degree in the community. Instead, we consider the least-degree neighbors for the anchor nodes in $\mathcal{A}^{large}$. We thus have the following definition.

\begin{definition}
\textbf{Anchor node (small degree) -- $\mathcal{A}^{small}$}: Suppose the $i$-th least-degree node in the set $X$ is $v^{X,s}_i$. Given a set of aligned large-degree nodes $\{u, v\} \in \mathcal{A}^{large}$, the aligned small-degree nodes can be defined as follows:

\begin{equation}
\mathcal{A}^{small} = \bigcup\limits^{\{u,v\} \in \mathcal{A}^{large}} \{v^{\mathcal{N}(u),s}_i,v^{\mathcal{N}(v),s}_j\}, i,j \in \{1,...,k_s\}, 
\end{equation}
where $k_s = \mathbf{Min}\{log_2|\mathcal{N}(u)|,log_2|\mathcal{N}(v)|\}$ and $\mathcal{N}(.)$ denotes the neighbor nodes.
\end{definition}

These least important nodes are trivial in the network (and have little influence on the network structure), therefore, their roles are basically the same and will not require a one-to-one relationship. Thus, the loss function for small-degree nodes is based on the average of the similarity of their combination.

\begin{equation}
\begin{split}
\mathcal{L}_{small} &= \sum\limits_{\{u,v\} \in \mathcal{A}^{small}} \frac{1}{k_s} (dist(\Phi^{a\rightarrow b}(\mathbb{E}_A(u)) , \mathbb{E}_B(v)) \\
&\quad + dist(\Phi^{b\rightarrow a}(\mathbb{E}_B(v)) , \mathbb{E}_A(u))).
\end{split}
\end{equation}

Network alignment models $\Phi^{a\rightarrow b}$ and $\Phi^{b\rightarrow a}$ will also be trained in an unsupervised manner to enhance the mapping accuracy. Given a node $u \in V_A$, the automapping embedding $\Phi^{b\rightarrow a}(\Phi^{a\rightarrow b}(\mathbb{E}_A(u)))$ should be as similar as the original embedding $\mathbb{E}_A(u)$, and the same for nodes $v \in V_B$. The automapping loss can be obtained as follows:

\begin{equation}
\begin{split}
\mathcal{L}_{us} &= dist(\Phi^{b\rightarrow a}(\Phi^{a\rightarrow b}(\mathbb{E}_A(u))), \mathbb{E}_A(u))\\
&\quad + dist(\Phi^{a\rightarrow b}(\Phi^{b\rightarrow a}(\mathbb{E}_B(v))), \mathbb{E}_B(v)).
\end{split}
\end{equation}

Given two networks of $G_A$ and $G_B$, the loss function for training the mappings $\Phi^{a\rightarrow b}$ and $\Phi^{b\rightarrow a}$ is obtained as follows:

\begin{equation}
\mathcal{L}_{alignment}\left(\mathbf{W}_{a, b}, \mathbf{b}_{a, b}, \mathbf{W}_{b, a}, \mathbf{b}_{b, a}\right) = \mathcal{L}_{large} + \mathcal{L}_{small} + \mathcal{L}_{us}.
\end{equation}
After optimizing $\mathcal{L}_{alignment}$, the trained mapping will align the remaining nodes between the two networks, facilitating the transfer of solutions between them.

To establish node-level alignment between two networks based on their learned embeddings, we first project the embeddings from one network into the latent space of the other using learned mapping functions. Specifically, for each node \( u \in V_A \), its embedding is transformed into the space of \( G_B \) using the mapping function \( \Phi^{A \rightarrow B} \). The node \( v \in V_B \) whose embedding is closest, in terms of Euclidean distance, to the mapped embedding of node \( u \) is identified, and its index is stored in the alignment mapping as \( M_{A \rightarrow B}(u) = v \).

\begin{algorithm}[t]
\caption{Network Alignment} 
\hspace*{0.02in} {\bf Input:} 
Networks $G_A = (V_A, E_A)$ and $G_B = (V_B, E_B)$\\
\hspace*{0.02in} {\bf Output:} 
Graph Mappings $\Phi^{a\rightarrow b}$ and $\Phi^{b\rightarrow a}$
\begin{algorithmic}[1]
\State  Obtain the embeddings of $V_A$ and $V_B$ as $\mathbb{E}_A = \mathbf{GAE}(G_A)$ and $\mathbb{E}_B = \mathbf{GAE}(G_B)$
\For{each epoch}
\State \texttt{\#Unsupervised learning where $u \in V_A$ and $v \in V_B$}
\State Map the embedding $\mathbb{E}_A(u)$ to the space of $G_B$ via $\Phi^{a\rightarrow b}(\mathbb{E}_A(u))$, and then map the embeddings of $G_B$ back to the space of $G_A$
\State Calculate the difference between $\Phi^{b\rightarrow a}(\Phi^{a\rightarrow b}(\mathbb{E}_A(u)))$ and $\mathbb{E}_A(u)$
\State Map the embedding $\mathbb{E}_B(v)$ to the space of $G_A$ via $\Phi^{b\rightarrow a}(\mathbb{E}_B(v))$, and then map the embeddings of $G_A$ back to the space of $G_B$
\State Calculate the difference between $\Phi^{a\rightarrow b}(\Phi^{b\rightarrow a}(\mathbb{E}_B(v)))$ and $\mathbb{E}_B(v)$
\State Calculate the unsupervised loss between the original embeddings and the automapped embeddings
\State \texttt{\#Supervised learning where $u \in V_A$ and $v \in V_B$, and $(u,v)$ is an anchor pair}
\State Map $u$ to the space of $G_B$ to obtain $\Phi^{a\rightarrow b}(\mathbb{E}_A(u))$, and calculating its difference with $\mathbb{E}_B(v)$
\State Map $v$ to the space of $G_A$ to obtain $\Phi^{b\rightarrow a}(\mathbb{E}_B(v))$, and calculating its difference with $\mathbb{E}_A(u)$
\State Use the anchor nodes to calculate the supervised loss
\EndFor
\State Update the parameters of mappings $\Phi^{a\rightarrow b}$ and $\Phi^{b\rightarrow a}$
\end{algorithmic}
\label{a4}
\end{algorithm}

\subsection{Many-network evolutionary optimization}
In evolutionary optimization for community deception, the chromosome is represented by a combination of genes (edges). The population of network $G_i \in \mathcal{G}$ can be defined as follows:

\begin{equation}
P_i = \left[\begin{array}{cccccc}
e_{1,1}^i & e_{1,2}^i & \cdots & \cdots & e_{1,\beta_i}^i & \rho_1^i\\
e_{2,1}^i & e_{2,2}^i & \cdots & \cdots & e_{2,\beta_i}^i & \rho_2^i\\
\vdots & \vdots & \vdots & \vdots & \vdots & \vdots \\
e_{n,1}^i & e_{n,2}^i & \cdots & \cdots & e_{n,\beta_i}^i & \rho_{n}^i
\end{array}\right],
\end{equation}
where the first $\rho$ edges are to add, and the rest $\beta-\rho$ are the edges to delete.

Let $\widetilde{\mathcal{C}_A} = \{\mathcal{C}_1, \mathcal{C}_2, \ldots, \mathcal{C}_k \}$ and $\widetilde{\mathcal{C}_B} = \{\mathcal{C}'_1, \mathcal{C}'_2, \ldots, \mathcal{C}'_{k'} \}$ represent two community partitions. The confusion matrix $m$ is employed to quantify the dissimilarity between these two community partitions. The entry $m_{ij}$ in the matrix denotes the number of common elements between $\mathcal{C}_i$ and $\mathcal{C}'_j$. To facilitate the disengagement of nodes from their initial community affiliations, we can formulate the fitness function as follows:
\begin{equation}
F = -\left( \sum_{i=1}^{|\widetilde{\mathcal{C}}|} \sum_{j=1}^{|\widetilde{\mathcal{C}}'|} \frac{|\mathcal{C}_i|}{|V|} \left( \frac{m_{ij}}{|\mathcal{C}_i|} \log_2 \frac{m_{ij}}{|\mathcal{C}_i|} \right) \right) * e^{-\max(m'_{ij})},
\end{equation}
where $m_{ij}$ denotes the number of common nodes between $\mathcal{C}_i$ and $\mathcal{C}'_j$, and $m'_{ij} = \frac{m_{ij}}{|\mathcal{C}^{'}_j|} * \log_2|\mathcal{C}^{'}_j|$.

{For a network $G \in \mathcal{G}$, considering the knowledge transferred from all the rest networks $\mathcal{G}/G$ is undesirable because 1) the optimization efficiency will be compromised with the increase in the networks participating in transfer; 2) the solution from networks that have a low similarity with the target network may bring negative transfer. Therefore, only the solutions from those networks similar to the target network will be transferred to the target population. Let $\mathcal{G}^\mathcal{S}_i$ be the set of top $\sqrt{|\mathcal{G}|}$ similar networks to network $G_i$, which satisfies}
\begin{equation}
{
\begin{cases}
|\mathcal{G}^\mathcal{S}_i| = \sqrt{|\mathcal{G}|}, \\
\forall G_j \in \mathcal{G}^\mathcal{S}_i, \forall G_k \notin \mathcal{G}^\mathcal{S}_i: \mathcal{S}^\mathcal{G}_{i, j} > \mathcal{S}^\mathcal{G}_{i,k}.
\end{cases}}
\end{equation}

The transferred solution requires several rounds of reproduction to be integrated into the population of the target domain. Combined with the interest of efficiency, we thus take $k=5$ empirically as the interval to determine whether to transfer or not. Let the best fitness values on $G_i$ in the current generation and $k$ generations ago be $F^i_{n_0}$ and $F^i_{n_k}$, respectively. Then, we can obtain the fitness difference between the current generation and $k$ generations ago and the fitness difference between $k$ and $2k$ generations ago as follows:

\begin{equation}
d_{1}^i=\left|F_{n_0}^i-F_{n_k}^i\right|, \quad
d_{2}^i=\left|F_{n_k}^i-F_{n_{2k}}^i\right|.
\end{equation}

Then, we define the transfer condition as follows:

\begin{equation}
\begin{cases}
d_{1}^i < d_{2}^i, \\
n \geq 2k, \\
n \bmod k = 0.
\end{cases}
\label{condition}
\end{equation}

The motivation of Eq. \ref{condition} is that if the evolution speed is slower than the previous, then excellent solutions of other similar domains will be transferred to facilitate the optimization.

To better use the transfer budget, we propose a self-adaptive mechanism to determine the number of solutions to transfer from $G \in \mathcal{G}^\mathcal{S}_i$ to $G_i$. Each transfer will be evaluated by its contribution to the new population by observing the difference between the current elite population $P^{Elite}_{i,n_0}$ and the elite population before the last transfer $P^{Elite}_{i,n_l}$, and its intersection with the last transfer ($T_{i\leftarrow j,n_l}$) from $G_j \in \mathcal{G}^\mathcal{S}_i$. That is,

\begin{equation}
I_{i\leftarrow j}^\mathcal{G}= \frac{|(P^{Elite}_{i,n_0} \setminus P^{Elite}_{i,n_l}) \bigcap T_{i\leftarrow j,n_l}|}{|P^{Elite}_{i,n_0}|},
\end{equation}
where $T_{i \leftarrow j} = \{ M_{i \leftarrow j}(s) \mid s \in S_j \},$
and \( S_j \) is a set of selected solutions from network \( G_j \), and \( M_{i \leftarrow j} \) is the function mapping each solution from the domain of \( G_j \) to that of \( G_i \).

{The similarity matrix is then updated as follows:}
\begin{equation}
\mathcal{S}^\mathcal{G}_{i,j} \leftarrow I_{i \leftarrow j}^\mathcal{G} + \mathcal{S}^\mathcal{G}_{i,j},
\label{Eq_simi}
\end{equation}
where $\mathcal{S}^\mathcal{G}_{i,j}$ refers to the similarity between $G_i$ and $G_j$, which can be used to control the ratio of transferred solutions from different networks, given a fixed number of transferred solutions $|T_{i}|$. The networks that are more similar to the target network should transfer more solutions, i.e.,

\begin{equation}
|T_{i \leftarrow j}| = |T_{i}| * \frac{\mathcal{S}^\mathcal{G}_{i,j}}{\sum\limits_{G_j \in \mathcal{G}^\mathcal{S}_i} \mathcal{S}^\mathcal{G}_{i,j}}.
\label{ea_number}
\end{equation}

{Elements in the similarity matrix corresponding to non-similar networks are set to zero ($\mathcal{S}^\mathcal{G}_{i,j}=0, G_j \notin \mathcal{G}^\mathcal{S}_i$) prior to optimization, whereas elements representing network similarity are normalized during each update.}

{
 
Let the population of network \( G_i \) be \( P_i \) and knowledge transfer occurs from similar networks \( G_j \in \mathcal{G}^\mathcal{S}_i \), and the transferred high-quality solution \( T_{i \leftarrow j} \) from a network \( G_j \in \mathcal{G}^\mathcal{S}_i \) is incorporated into the population to assist the optimization process. Therefore, the updated population is given by:
}

\begin{equation}
{
P_i =
\begin{cases}
P_i \cup \bigcup_{G_j \in \mathcal{G}^\mathcal{S}_i} T_{i \leftarrow j}, & \text{if knowledge transfer occurs,} \\
P_i, & \text{otherwise.}
\end{cases}
}
\label{Trans_pop}
\end{equation}

To better utilize the knowledge of other domains, we introduce a new mutation mechanism that leverages elite edges from other networks to replace edges in the current domain. The probability of selecting a network \( G_j \in \mathcal{G}^\mathcal{S}_i \) for knowledge transfer is defined as follows:

\begin{equation}
\mathbb{P}^{M}_{i \leftarrow j} = \frac{\mathcal{S}^\mathcal{G}_{i,j}}{\sum\limits_{G_j \in \mathcal{G}^\mathcal{S}_i} \mathcal{S}^\mathcal{G}_{i,j}}.
\label{mutate}
\end{equation}

Given a network \( \mathcal{G}_i \), the deletion and addition candidates are denoted by \( E_i^{M,D} \) and \( E_i^{M,A} \), respectively. When knowledge transfer is applied, these candidates are derived from the transferred solutions of assisted networks \( G_j \in \mathcal{G}^\mathcal{S}_i \), which are organized into addition (\( T_{i \leftarrow j}^A \)) and deletion (\( T_{i \leftarrow j}^D \)) sets. The mutation candidate sets are then updated accordingly as follows:

\begin{equation}
{
E_i^{M,A} = \bigcup_{G_j \in \mathcal{G}^\mathcal{S}_i} T_{i \leftarrow j}^D, \quad E_i^{M,D} =  \bigcup_{G_j \in \mathcal{G}^\mathcal{S}_i} T_{i \leftarrow j}^A,
}
\end{equation}
where \( T_{i \leftarrow j}^A \) represents the set of edges for addition, and \( T_{i \leftarrow j}^D \) represents the set of edges for deletion from the transferred solution of network \( G_j \in \mathcal{G}^\mathcal{S}_i \).

As for the selection in our work, each individual has an equal probability of being selected, promoting diversity and helping prevent premature convergence on suboptimal solutions. Additionally, we adopt an elitist mechanism to ensure that the highest-quality solutions are preserved.

As different networks have different budgets $\beta$, the repairs to transferred solutions are necessary to ensure compatibility when the sizes of the transferred solution and the solution of the target network are not matched. Given an assisted network $G_A$ and a target network $G_T$, if $\beta_A > \beta_T$, then the edges in the transferred solution will be removed randomly until the length of the transferred solution meets the requirement. Otherwise, the edges will be sampled from existing edges as deletion and nonexistent edges as addition to make the transferred solution valid. {The pseudocode of the many-network evolutionary optimization can be found in Algorithm \ref{a5}.}




\begin{algorithm}[t]
\caption{Many-Network Evolutionary Optimization} 
\hspace*{0.02in} {\bf Input:} 
Networks $G_i=(V_i, E_i) \in \mathcal{G}$\\
\hspace*{0.02in} {\bf Output:} 
The optimal solutions for the task across each network
\begin{algorithmic}[1]
\State  Initialize population for each network 
\State  Initialize similarity matrix 
\State Identify the top-k similar networks for each network
\While{the iteration does not reach the limit}
\For{network $G_i \in \mathcal{G}$} 
\State Perform selection 
\If{condition meets} 
\For{network $G_j$ in $\mathcal{G}^\mathcal{S}_i$}
\State Compute the number of solutions transferred from $G_j$
\State Transfer elitism solutions from $G_j$ and add them to the population $P_i$ 
\EndFor
\State Update and normalize similarity matrix
\EndIf
\State Perform crossover operation
\State Perform mutation operation
\EndFor
\EndWhile
\end{algorithmic}
\label{a5}
\end{algorithm}

{\subsection{Complexity analysis}}
{In the proposed MDEO, the process begins with the calculation of graph similarity where this step requires $O(|\widetilde{\mathcal{C}}| \cdot |\widetilde{\mathcal{C'}}|)$ operations for each pair of graphs}. {Following the similarity calculations, generating embeddings using the GAE is with the complexity of $O(|E|\cdot d)$. The training of the network alignment model incorporates both unsupervised and supervised machine learning techniques, requiring the time of $O(|V| \cdot d)$ and $O(\mathcal{A} \cdot d)$ respectively for each epoch where $\mathcal{A}$ is the number of anchor pair and $d$ is the dimension of learned embeddings. }

{If the networks are optimized independently, the time complexity is $O(|\mathcal{G}| \cdot |P|  \cdot N_t)$, where $N_t$ denotes the number of evolutionary rounds, and $|P|$ is the population size in each network. In the MDEO framework, the solutions from $\log |\mathcal{G}|$ networks are transferred to assist others, but the number of solutions transferred, denoted by $|T|$, is fixed, and all assisted networks share this budget. Consequently, the worst-case complexity for MDEO could be $O(|\mathcal{G}| \cdot |P \cup T| \cdot N_t)$. However, transfers only occur at least every five generations and additionally must satisfy the requirements of Eq. \ref{condition}. Therefore, the practical complexity will be lower than this worst-case scenario. Since $|T|$ is a constant and significantly smaller than $|P|$, the complexity of MDEO will only be marginally higher than that of separately optimized networks, i.e., single-domain evolutionary optimization (SDEO), thereby justifying the practical efficiency of MDEO.}\\

\section{Experimental studies}\label{experiment}
In this section, eight real-world networks are utilized to examine our proposed MDEO, and the parameter sensitivity and structural change are also investigated.

\subsection{Experimental setting}
To obtain graph embeddings, we adopt a two-layer GCN as the encoder where the degree centrality, closeness centrality and community information are collected as node features.
In the process of evolution, the population size of each network is set to 100, and the probabilities of crossover and mutation are set to 0.5 and 0.1, respectively. The number of transferred solutions is 30. {In our study, we use two classical community detection algorithms, FastGreedy \cite{clauset2004finding} and WalkTrap \cite{pons2005computing}, as attackers to assess whether the community structure has been effectively obfuscated through community deception methods. The results are averaged over 20 independent runs and the implementation data is available online\footnote{https://networkrepository.com/}.}

\subsection{{Benchmark}}
{The mainstream approaches to community deception include heuristic methods, GNN-based method, and evolutionary optimization techniques. Accordingly, some of each have been chosen as benchmarks, specifically:} {\textbf{RAM}: A heuristic approach that randomly rewires all possible edges in the network \cite{waniek2018hiding};} {\textbf{DICE}: A heuristic approach that randomly deletes intra-community edges and adds inter-community edges \cite{waniek2018hiding};}
{\textbf{Mod}: This method uses the genetic algorithm to optimize modularity to modify the network edges to conceal community information \cite{chen2019ga};} {\textbf{CD-ATTACK}: This method alters links by using constrained adversarial graphs generator based on graphs neural networks (GNNs) \cite{li2020adversarial};}
{\textbf{CoeCo}: This approach employs a divide-and-conquer strategy for scalable evolutionary optimization. It optimizes both the subgraphs and the original graph cooperatively to identify the optimal edge set for community obfuscation \cite{zhao2023Obfuscating};} {\textbf{SDEO}: The classical evolutionary optimization without knowledge transfer, where each network is optimized separately, providing a baseline for comparison against our proposed MDEO.}



\subsection{Experiment on real-world networks}
To examine the efficacy of different community deception algorithms, we adopt two metrics of normalized mutual information (NMI) \cite{amelio2015normalized} and adjusted rand index (ARI) \cite{vinh2009information} to observe the change in community structure, shown in Table \ref{NMI}. The lower the NMI and ARI are, the better protection the community deception algorithm achieves. The statistical disparities between MDEO and other methodologies were assessed using the Wilcoxon rank sum test, setting the significance level at \( p=0.05 \). Symbols `+' and `-' are used to show whether the alternative methods outperform or underperform compared to MDEO, respectively, and `$\approx$' signals performance equivalence. The community deception algorithms are tested with two attackers (community detection algorithms) on eight networks of different domains. {As observed, our proposed MDEO consistently outperforms other methods, including heuristic approaches (RANDOM, DICE), metaheuristic algorithms (MoD, CoeCo), and the GNN-based method (CD-ATTACK). It also demonstrates superior performance compared to SDEO, as indicated by its lower average NMI and ARI scores across the evaluated networks.} This result indicates the effectiveness of our method in securing community information. On the other hand, the result is subject to the attacker. For example, the NMI and ARI in Adjnoun when dealing with the attack from FastGreedy, are lower than other networks. Network is also an important variable, as suggested by that the NMI and ARI in Adjnoun are higher when dealing with WalkTrap than other networks. 

\begin{table*}
\caption{{The comparison of the mean and standard deviation of NMI and ARI obtained from different community deception algorithms on eight tested networks.  FastGreedy and WalkTrap are used to examine the difference in community structure before and after modification, respectively.}}
\centering
\subtable[FastGreedy]{
\resizebox{182mm}{14mm}{
\begin{tabular}{cccccccccccccccc}  
\Xhline{5\arrayrulewidth} 
\multirow{2}{*}{\textbf{Network}} &  
\multicolumn{7}{c}{\textbf{NMI}} & \multicolumn{7}{c}{\textbf{ARI}} \\
\cmidrule(lr){2-8} \cmidrule(lr){9-15} 
& \textbf{RAM} & \textbf{DICE} & \textbf{Mod} & {\textbf{CD-ATTACK}} & \textbf{CoeCo} & \textbf{SDEO} & \textbf{MDEO} 
& \textbf{RAM} & \textbf{DICE} & \textbf{Mod} & {\textbf{CD-ATTACK}} & \textbf{CoeCo} & \textbf{SDEO} & \textbf{MDEO} \\
\midrule  
\textbf{Dolphins}&0.75$\pm$0.06(-)&0.74$\pm$0.08(-)&0.76$\pm$0.09(-)&{0.72$\pm$0.02(-)}&{0.51$\pm$0.03(-)}&0.51$\pm$0.05(-)&\cellcolor{gray!40}0.43$\pm$0.04&0.70$\pm$0.07(-)&0.69$\pm$0.09(-)&0.71$\pm$0.11(-)&{0.54$\pm$0.03(-)}&{0.35$\pm$0.04(-)}&0.31$\pm$0.05(-)&\cellcolor{gray!40}{0.24}$\pm$0.03\cr
\textbf{Lesmis}&0.79$\pm$0.11(-)&0.75$\pm$0.12(-)&0.80$\pm$0.09(-)&{0.79$\pm$0.08(-)}&{0.57$\pm$0.05(-)}&0.55$\pm$0.02(-)&\cellcolor{gray!40}{0.47}$\pm$0.04&0.72$\pm$0.18(-)&0.65$\pm$0.19(-)&0.76$\pm$0.11(-)&{0.69$\pm$0.11(-)}&{0.35$\pm$0.03(-)}&0.31$\pm$0.03(-)&\cellcolor{gray!40}{0.25}$\pm$0.02\cr 
\textbf{Polbooks}&0.85$\pm$0.10(-)&0.89$\pm$0.08(-)&0.92$\pm$0.06(-)&{0.75$\pm$0.06(-)}&{0.59$\pm$0.04(-)}&0.54$\pm$0.05($\approx$)&\cellcolor{gray!40}{0.52}$\pm$0.04&0.84$\pm$0.12(-)&0.89$\pm$0.09(-)&0.92$\pm$0.079(-)&{0.66$\pm$0.11(-)}&{0.54$\pm$0.05(-)}&0.45$\pm$0.07(-)&\cellcolor{gray!40}{0.41}$\pm$0.05\cr
\textbf{Adjnoun}&0.54$\pm$0.12(-)&0.42$\pm$0.09(-)&0.48$\pm$0.12(-)&{0.55$\pm$0.06(-)}&{0.25$\pm$0.03(-)}&0.24$\pm$0.04(-)&\cellcolor{gray!40}{0.24}$\pm$0.03&0.43$\pm$0.16(-)&0.29$\pm$0.12(-)&0.37$\pm$0.15(-)&{0.38$\pm$0.09(-)}&{0.08$\pm$0.02($\approx$)}&0.09$\pm$0.02(-)&\cellcolor{gray!40}{0.07}$\pm$0.02\cr
\textbf{Erdos}&0.68$\pm$0.09(-)&0.67$\pm$0.07(-)&0.70$\pm$0.08(-)&{0.73$\pm$0.04(-)}&{\cellcolor{gray!40}{0.47$\pm$0.02(+)}}&0.50$\pm$0.03(-)&{0.49}$\pm$0.02&0.58$\pm$0.13(-)&0.58$\pm$0.10(-)&0.61$\pm$0.13(-)&{0.61$\pm$0.07(-)}&{0.28$\pm$0.02($\approx$)}&0.28$\pm$0.03($\approx$)&\cellcolor{gray!40}{0.27}$\pm$0.02\cr
\textbf{USAir}&0.73$\pm$0.11(-)&0.65$\pm$0.13(-)&0.73$\pm$0.11(-)&{0.63$\pm$0.06(-)}&{0.43$\pm$0.05(-)}&0.48$\pm$0.057(-)&\cellcolor{gray!40}{0.41}$\pm$0.05&0.70$\pm$0.17(-)&0.64$\pm$0.16(-)&0.74$\pm$0.14(-)&{0.53$\pm$0.09(-)}&{0.29$\pm$0.05(-)}&0.34$\pm$0.06(-)&\cellcolor{gray!40}{0.27}$\pm$0.05\cr
\textbf{Netscience}&0.95$\pm$0.03(-)&0.93$\pm$0.03(-)&0.96$\pm$0.03(-)&{0.91$\pm$0.01(-)}&{0.82$\pm$0.02(-)}&0.83$\pm$0.02(-)&\cellcolor{gray!40}{0.80}$\pm$0.02&0.90$\pm$0.07(-)&0.85$\pm$0.085(-)&0.91$\pm$0.07(-)&{0.78$\pm$0.05(-)}&{0.57$\pm$0.03(-)}&0.57$\pm$0.03(-)&\cellcolor{gray!40}{0.53}$\pm$0.04\cr
\textbf{BioCelegans}&0.52$\pm$0.06(-)&0.49$\pm$0.04(-)&0.49$\pm$0.07(-)&{0.53$\pm$0.05(-)}&{0.41$\pm$0.03(-)}&0.40$\pm$0.04($\approx$)&\cellcolor{gray!40}{0.39}$\pm$0.03&0.42$\pm$0.08(-)&0.37$\pm$0.07(-)&0.39$\pm$0.01(-)&{0.38$\pm$0.06(-)}&{0.25$\pm$0.03(-)}&0.22$\pm$0.03($\approx$)&\cellcolor{gray!40}{0.21}$\pm$0.02\cr
\Xhline{5\arrayrulewidth} 
\end{tabular} } }

\subtable[WalkTrap]{
\resizebox{182mm}{14mm}{
\begin{tabular}{cccccccccccccccc}  
\Xhline{5\arrayrulewidth} 
\multirow{2}{*}{\textbf{Network}} &  
\multicolumn{7}{c}{\textbf{NMI}} & \multicolumn{7}{c}{\textbf{ARI}} \\
\cmidrule(lr){2-8} \cmidrule(lr){9-15} 
& \textbf{RAM} & \textbf{DICE} & \textbf{Mod} & {\textbf{CD-ATTACK}} & \textbf{CoeCo} & \textbf{SDEO} & \textbf{MDEO} 
& \textbf{RAM} & \textbf{DICE} & \textbf{Mod} & {\textbf{CD-ATTACK}} & \textbf{CoeCo} & \textbf{SDEO} & \textbf{MDEO} \\
\midrule  
\textbf{Dolphins}&0.67$\pm$0.06(-)&0.64$\pm$0.08(-)&0.67$\pm$0.06(-)&{{0.55$\pm$0.19(-)}}&{0.58$\pm$0.03(-)}&0.57$\pm$0.03(-)&\cellcolor{gray!40}{0.54}$\pm$0.04&0.51$\pm$0.09(-)&0.48$\pm$0.08(-)&0.51$\pm$0.07(-)&{{0.39$\pm$0.18(-)}}&{0.33$\pm$0.03(-)}&0.30$\pm$0.03(-)&\cellcolor{gray!40}{0.27}$\pm$0.03\cr 
\textbf{Lesmis}&0.90$\pm$0.04(-)&0.86$\pm$0.05(-)&0.90$\pm$0.04(-)&{{0.76$\pm$0.05(-)}}&{0.78$\pm$0.02(-)}&0.81$\pm$0.02(-)&\cellcolor{gray!40}{0.74}$\pm$0.02&0.80$\pm$0.10(-)&0.72$\pm$0.11(-)&0.77$\pm$0.12(-)&{{0.53$\pm$0.12(-)}}&{0.56$\pm$0.03(-)}&0.55$\pm$0.03(-)&\cellcolor{gray!40}{0.48}$\pm$0.02\cr 
\textbf{Polbooks}&0.92$\pm$0.06(-)&0.88$\pm$0.04(-)&0.90$\pm$0.07(-)&{{0.87$\pm$0.05(-)}}&{0.67$\pm$0.02(-)}&0.64$\pm$0.02($\approx$)&\cellcolor{gray!40}{0.63}$\pm$0.01&0.93$\pm$0.06(-)&0.88$\pm$0.06(-)&0.89$\pm$0.12(-)&{{0.90$\pm$0.05(-)}}&{0.51$\pm$0.02(-)}&0.49$\pm$0.02($\approx$)&\cellcolor{gray!40}{0.48}$\pm$0.02\cr
\textbf{Adjnoun}&0.79$\pm$0.04(-)&0.76$\pm$0.05(-)&0.80$\pm$0.04(-)&{{0.74$\pm$0.03(-)}}&{0.71$\pm$0.02(-)}&\cellcolor{gray!40}0.72$\pm$0.02(+)&0.72$\pm$0.02&0.52$\pm$0.11(-)&0.48$\pm$0.15(-)&0.52$\pm$0.13(-)&{{0.42$\pm$0.09(-)}}&{0.22$\pm$0.03(-)}&{0.20}$\pm$0.04(+)&\cellcolor{gray!40}{0.20}$\pm$0.02\cr 
\textbf{Erdos}&0.90$\pm$0.01(-)&0.87$\pm$0.02(-)&0.88$\pm$0.02(-)&{{0.68$\pm$0.10(-)}}&{0.85$\pm$0.02(-)}&0.85$\pm$0.01(+)&\cellcolor{gray!40}{0.85}$\pm$0.01&0.70$\pm$0.06(-)&0.62$\pm$0.07(-)&0.63$\pm$0.07(-)&{{0.55$\pm$0.01(-)}}&{0.50$\pm$0.04(-)}&0.47$\pm$0.04($\approx$)&\cellcolor{gray!40}{0.45}$\pm$0.03\cr
\textbf{USAir}&0.83$\pm$0.05(-)&0.82$\pm$0.06(-)&0.83$\pm$0.04(-)&{\cellcolor{gray!40}{0.66$\pm$0.01(+)}}&{0.77$\pm$0.02(-)}&0.77$\pm$0.01(-)&{0.75}$\pm$0.01&0.77$\pm$0.07(-)&0.74$\pm$0.14(-)&0.76$\pm$0.10(-)&{\cellcolor{gray!40}{0.34$\pm$0.01(+)}}&{0.47$\pm$0.07(-)}&0.48$\pm$0.04(-)&{0.44}$\pm$0.04\cr
\textbf{Netscience}&0.94$\pm$0.03(-)&0.92$\pm$0.02(-)&0.93$\pm$0.02(-)&{0.89$\pm$0.02(-)}&{0.88$\pm$0.01(-)}&0.86$\pm$0.01(-)&\cellcolor{gray!40}{0.85}$\pm$0.01&0.82$\pm$0.11(-)&0.75$\pm$0.10(-)&0.80$\pm$0.10(-)&{0.68$\pm$0.07(-)}&{0.62$\pm$0.02(-)}&0.53$\pm$0.03(-)&\cellcolor{gray!40}{0.51}$\pm$0.03\cr
\textbf{BioCelegans}&0.78$\pm$0.03(-)&0.72$\pm$0.03(-)&0.79$\pm$0.03(-)&{0.72$\pm$0.02(-)}&{0.73$\pm$0.01(-)}&{0.74$\pm$0.01(-)}&\cellcolor{gray!40}{0.73}$\pm$0.01&0.45$\pm$0.06(-)&0.50$\pm$0.06(-)&0.47$\pm$0.08(-)&{0.35$\pm$0.04(-)}&{0.31$\pm$0.03(-)}&0.31$\pm$0.02(-)&\cellcolor{gray!40}{0.30}$\pm$0.02\cr
\Xhline{5\arrayrulewidth} 
\end{tabular} } }

\label{NMI}
\end{table*}

To examine the effectiveness of our proposed MDEO, we compare it with the traditional optimization SDEO by observing their evolution process. As observed in Figures \ref{Ablation}a and \ref{Ablation}b, SDEO converges around or before $50th$ generation while MDEO still exhibits an upward trend even in $200th$ generation on all networks, indicating that MDEO can effectively enhance the optimization and is not easily trapped into local optimality. It is worth noting that the optimization curves are ladder-shaped, meaning the solution transferred from other similar networks can improve the optimization on the target network. All curves depict an upward trend as shown in Figures \ref{Ablation}a and \ref{Ablation}b, depicting the high robustness of MDEO.

\begin{figure*}
        \centering
	\ContinuedFloat*
	\includegraphics[height=8cm,width=15cm]{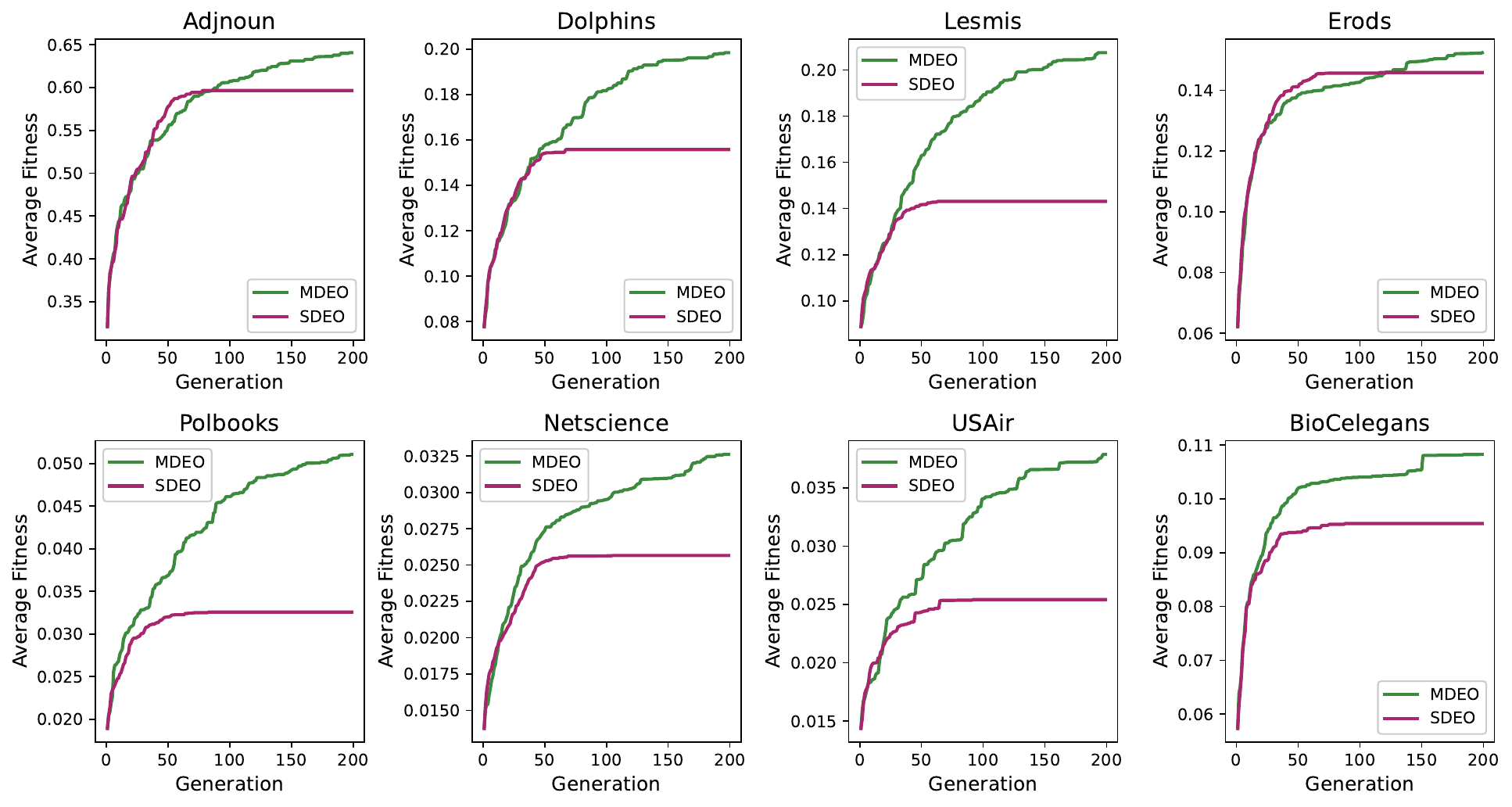}
	\caption{{The optimization curve of MDEO and SDEO on eight tested networks for the illustrative task of community deception, with FastGreedy as the attacker. }}
\end{figure*}

\begin{figure*}
    \centering
	\ContinuedFloat
	\includegraphics[height=8cm,width=15cm]{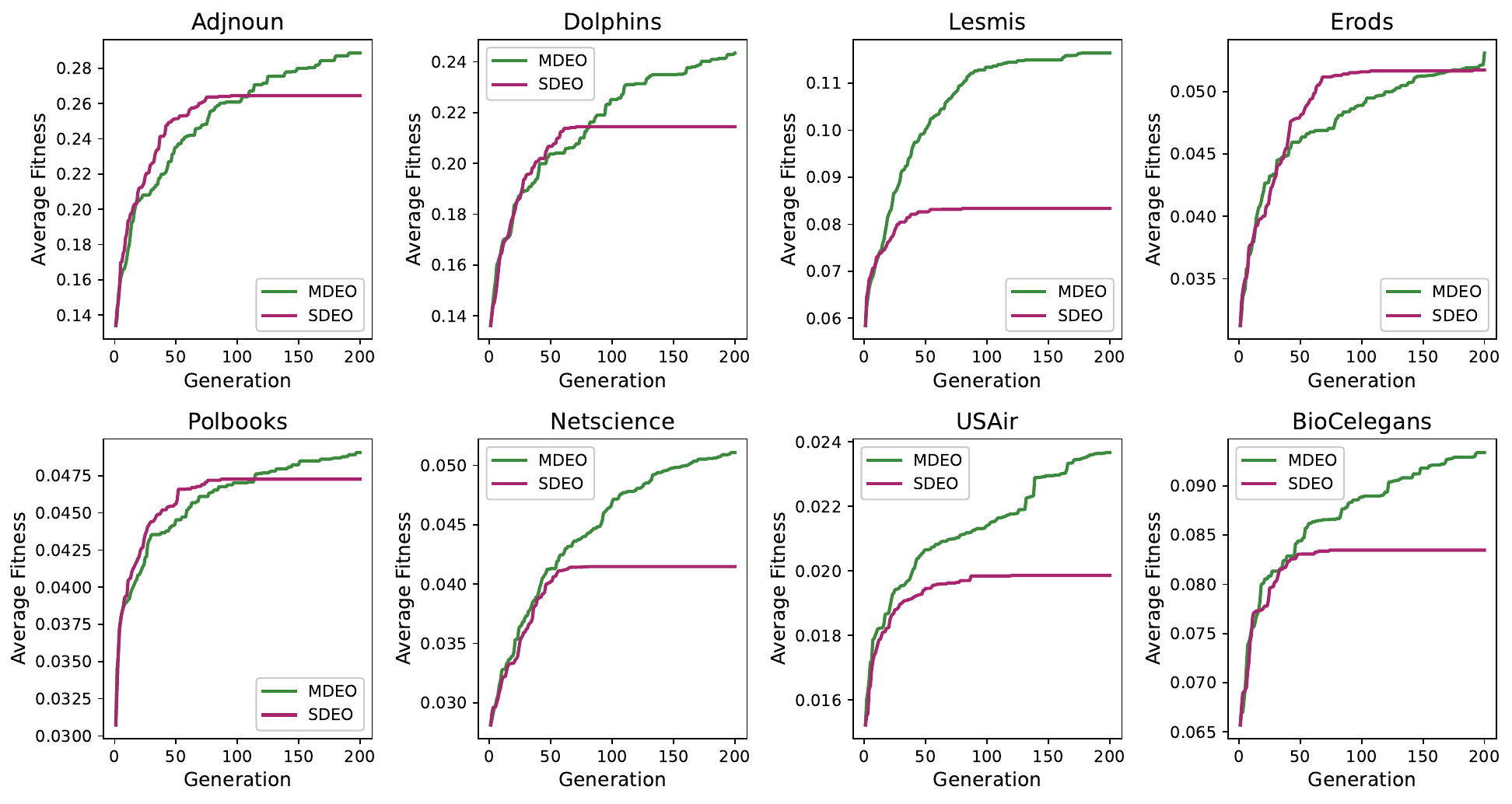}
	\caption{{The optimization curve of MDEO and SDEO on eight tested networks for the illustrative task of community deception, with WalkTrap as the attacker. }}
\label{Ablation}
\end{figure*}

\subsection{Parameter sensitivity analysis}



{In this section, we investigate the parameter sensitivity of the MDEO by varying three key parameters in Table \ref{parameter_analysis}:  The transfer interval ($\text{k}$), the number of transfer solutions ($|\text{T}|$), and the number of assisted networks ($|\mathcal{G}^S|$), with the default configuration set to $\text{k} = 5$, $|\text{T}| = 30$, and $|\mathcal{G}^S| = 3$. Compared to the default setting, increasing the transfer interval (e.g., $\text{k} = 10$ and $\text{k} = 20$) generally results in slightly degraded performance across most networks, indicating that more frequent knowledge transfer (lower $\text{k}$) is beneficial for optimization. Likewise, reducing the number of transfer solutions (e.g., $|\text{T}| = 10$ and $|\text{T}| = 20$) leads to small drops in performance when compared with the default $|\text{T}| = 30$, suggesting that transferring more solutions helps guide the search more effectively. Furthermore, reducing the number of assisted networks (i.e., $|\mathcal{G}^S| = 1$ or $2$) also yields worse results relative to the default, reinforcing the idea that leveraging more source networks provides richer and more useful guidance. We also investigate the role of community partitioning in network alignment by comparing community-based alignment with direct alignment of high- and low-degree nodes. The results show that community-based alignment more effectively facilitates knowledge transfer.} The results clearly support that more frequent and comprehensive knowledge transfer leads to better optimization outcomes, indicating the effectiveness of the proposed MDEO.

\begin{table*}
\caption{{Parameter sensitivity analysis of MDEO with the knowledge transfer interval $\text{k}$ set to 5, the number of transfer solutions $|\text{T}|$ set to 30, and the number of assisted networks $|\mathcal{G}^\mathcal{S}|$ set to 3. MDEO incorporates community-based alignment by default, referred to CA. The tested attacker is the FastGreedy algorithm. The shadow cell is a better setting than default.}}
\centering
\begin{tabular}{c|c|cc|cc|cc|c}  
\Xhline{5\arrayrulewidth} 
\textbf{Network} & \textbf{W/O CA} & $\mathbf{k=10}$ & $\mathbf{k=20}$ & $\mathbf{|T| = 10}$ & $\mathbf{|T|=20}$ & $\boldsymbol{|\mathcal{G}^\mathcal{S}|=1}$ & $\boldsymbol{|\mathcal{G}^\mathcal{S}|=2}$ & \textbf{Default} \\
\midrule  
\textbf{Dolphins}&{0.189}&0.187&0.182&0.188&0.189&0.167&{0.188}&{0.198}\cr
\textbf{Lesmis}&\cellcolor{gray!40}{0.213}&0.195&0.186&0.182&0.197&0.205&0.207&{0.208}\cr 
\textbf{Polbooks}&0.044&0.048&0.041&0.038&0.047&0.044&{0.051}&0.051\cr
\textbf{Adjnoun}&0.644&0.623&0.621&0.578&0.633&0.644&\cellcolor{gray!40}{0.653}&0.649\cr 
\textbf{Erdos}&0.151&0.149&0.150&0.146&0.149&0.141&0.148&{0.152}\cr
\textbf{USAir}&0.029&0.032&0.028&0.028&0.037&0.030&0.037&{0.039}\cr
\textbf{Netscience}&0.028&0.030&0.027&0.030&0.032&0.029&0.031&{0.032}\cr
\textbf{BioCelegans}&0.104&0.106&0.104&0.105&0.106&0.101&{0.105}&0.108\cr
\hline
\textbf{Comparison} &{1/7}&{0/8}&{0/8}&{0/8}&{0/8}&{0/8}&{1/7}&-\cr
\Xhline{5\arrayrulewidth} 
\end{tabular} 
\label{parameter_analysis}
\end{table*}

\begin{figure}
\centering
\includegraphics[height=4.5cm,width=8.8cm]{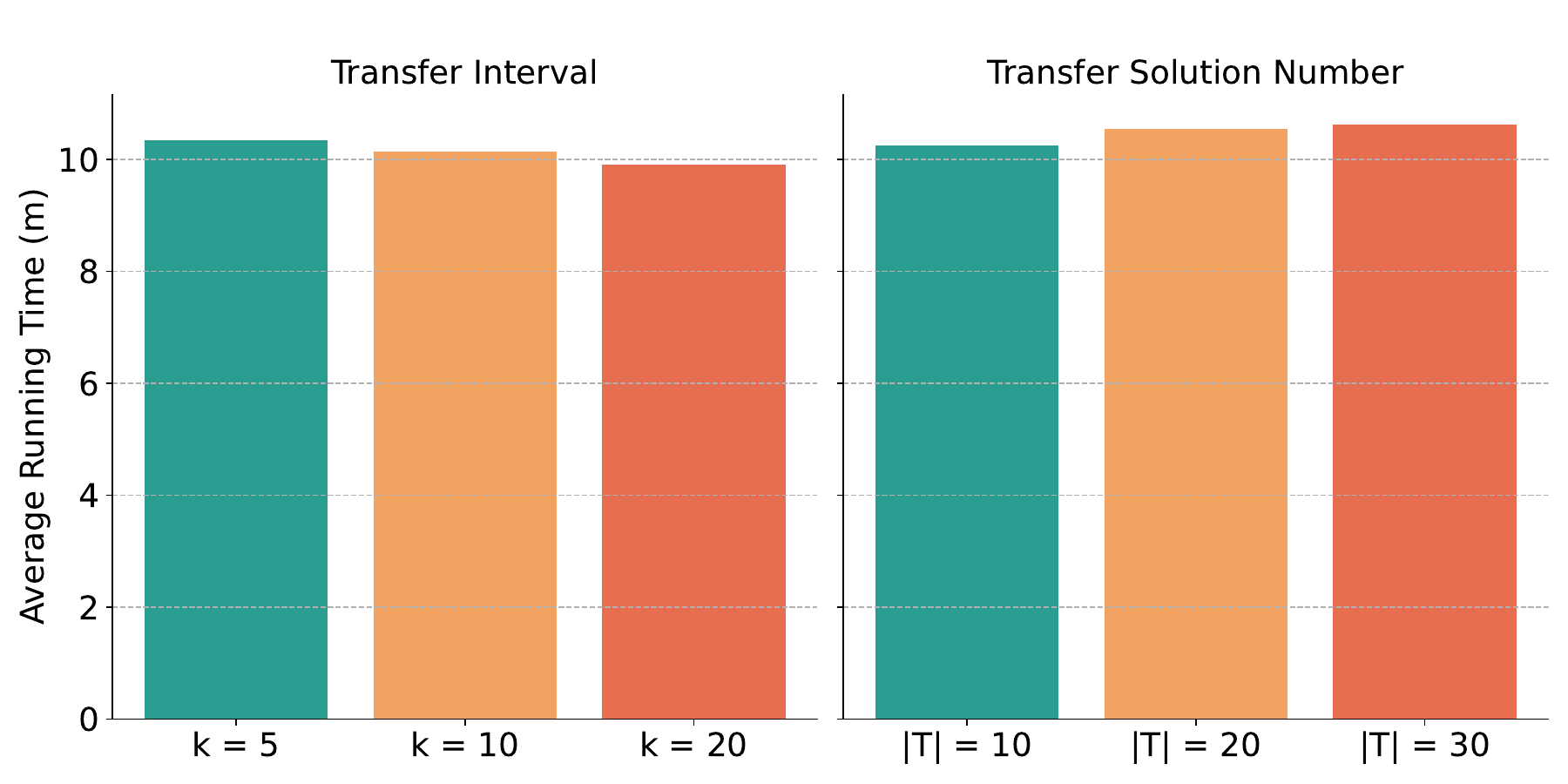}
\caption{{Average running time of MDEO under different parameter settings when optimizing eight real-world networks simultaneously.}}
\label{running_time}
\end{figure}

{As for the efficiency, Figure~\ref{running_time} illustrates the average running time of the MDEO for different parameter configurations related to knowledge transfer. The left subplot examines the impact of varying the transfer interval ($\text{k}$) while the right subplot evaluates different transfer solution numbers ($|\text{T}|$). The results indicate that both parameters have a marginal influence on the average running time. Specifically, as the transfer interval increases, a slight decrease in running time is observed, suggesting reduced overhead from less frequent transfers. Conversely, increasing the number of transfer solutions slightly increases the running time, likely due to the additional computational cost of processing more solutions. As observed, the impact of knowledge transfer settings on running time is minimal, indicating that the MDEO achieves an excellent balance between effectiveness and efficiency.}

\subsection{Structural change}
The structural change in some indices after perturbation is shown in Table \ref{preservation}, through which we can observe which kind of edges that the community deception algorithm prefers to modify. CC refers to the clustering coefficient, measuring the extent to which nodes tend to cluster together. The clustering coefficients of all networks have decreased, meaning the local connectivity within the network is becoming lower. On the other hand, the average shortest distance (ASD) gets improved after rewiring edges, which suggests the distance between communities is shortened. As for the centralities, the change in Betweenness is universally larger than the change in PageRank, indicating the proposed community deception algorithm tends to change the bridge edges. The observed reduction in modularity indicates that our algorithm effectively weakens the internal connections of individual communities while simultaneously enhancing their external connectivity.

\begin{table*}
\caption{The structural modifications resulting from edge alterations performed by MDEO, with FastGreedy as the testing attacker.}
\centering
\begin{tabular}{lcccccc}
\Xhline{5\arrayrulewidth}  Data & $|E^+\cup E^-|$ & CC & ASD & $20 \%-$ Betweenness  & $20 \%-$ PageRank & Modularity \\
\hline Dolphins & 10 & $0.309 \rightarrow 0.291$ & $3.357 \rightarrow 3.051$ & $1 \rightarrow 0.667$ & $1 \rightarrow 0.933$ & $0.495 \rightarrow 0.442$ \\
Lesmis & 10 & $0.498 \rightarrow 0.490$ & $2.641 \rightarrow 2.582$ & $1 \rightarrow 0.913$ & $1 \rightarrow 0.916$ & $0.500 \rightarrow 0.489$ \\
Polbooks & 20 & $0.348 \rightarrow 0.334$ & $3.079 \rightarrow 2.830$ & $1 \rightarrow 0.702$ & $1 \rightarrow 0.974$ & $0.502 \rightarrow 0.472$ \\
Adjnoun & 20 & $0.157 \rightarrow 0.156$ & $2.536 \rightarrow 2.523$ & $1 \rightarrow 0.923$ & $1 \rightarrow 0.975$ & $0.294 \rightarrow 0.272$ \\
Netscience & 30 & $0.431 \rightarrow 0.413$ & $6.042 \rightarrow 5.322$ & $1 \rightarrow 0.844$ & $1 \rightarrow 0.946$ & $0.839 \rightarrow 0.798$ \\
Erdos & 50 & $0.214 \rightarrow 0.206$ & $4.021 \rightarrow 3.958$ & $1 \rightarrow 0.932$ & $1 \rightarrow 0.958$ & $0.513 \rightarrow 0.484$ \\
BioCelegans & 50 & $0.124 \rightarrow 0.123$ & $2.664 \rightarrow 2.664$ & $1 \rightarrow 0.948$ & $1 \rightarrow 0.959$ & $0.395 \rightarrow 0.390$ \\
USAir & 100 & $0.396 \rightarrow 0.386$ & $2.738 \rightarrow 2.715$ & $1 \rightarrow 0.902$ & $1 \rightarrow 0.973$ & $0.319 \rightarrow 0.303$ \\
\Xhline{5\arrayrulewidth} 
\end{tabular}
\label{preservation}
\end{table*}

\subsection{{Generalizability analysis}}\label{Sec.ge}

{
The MDEO framework, originally demonstrated through the edge-level task of community deception, is fundamentally designed for generalizability. It can be effectively applied to a wide range of graph-structured combinatorial problems, encompassing both node-level and edge-level objectives. By appropriately adapting the solution representation and fitness function to suit the specific optimization goal, MDEO is capable of addressing diverse challenges across various domains.
}

{
As a way of illustration, we test the classical node-level task, i.e., the influence maximization task~\cite{jiang2011simulated}, where the solution sizes are set to 10 for the Email and Polblogs networks, and 15 for the Facebook and Wiki networks. As shown in Figures~\ref{IM}, MDEO consistently achieves higher average fitness values compared to SDEO, highlighting its effectiveness across different network structures.  These representative tasks at each level collectively demonstrate the method's generality and adaptability.



\begin{figure}
\centering
\includegraphics[height=7cm,width=7cm]{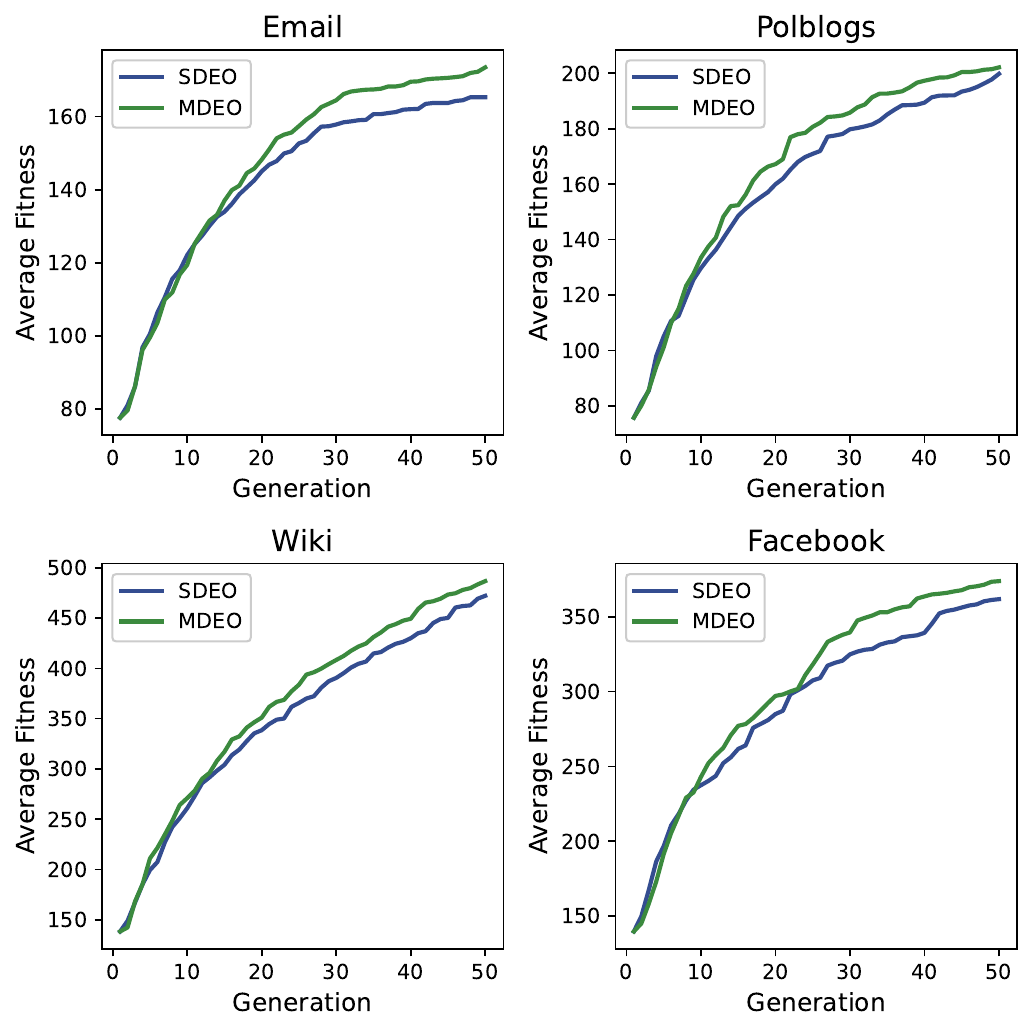}
\caption{{The optimization curve of MDEO and SDEO in addressing the node-level problem of influence maximization.}}
\label{IM}
\end{figure}

\section{Discussion}\label{sec.discussion}
{Although our MDEO framework has demonstrated promising performance, there still exists some challenges. Many real-world networks vary considerably in scale, and direct knowledge transfer across such imbalanced scenarios may introduce negative transfer effects. To address this issue, we observe that certain fundamental structural properties, such as modularity, centrality, and connectivity patterns, tend to remain consistent across different scales, providing a reliable foundation for effective knowledge transfer between networks of varying sizes. Building on this insight, we will explore extending MDEO to support a bidirectional transfer mechanism that enables adaptive knowledge exchange between large and small networks. This enhancement is designed to improve the robustness and adaptability of our approach in imbalanced-size scenarios, thereby enhancing its effectiveness across heterogeneous domains. Developing this bidirectional optimization framework is a key direction for future work and is expected to expand the practical applicability of MDEO.}

\section{Conclusion}\label{conclusion}
In this work, we have explored a framework--multi-domain evolutionary optimization (MDEO). We actualized this concept within the realm of network structures, predicated on the inherent shared properties of real-world networks. The proposed method was validated on the task of community deception with eight different real-world networks of varying sizes from various domains, and the experimental results show that the fitness values of evolutionary optimization on different networks have been improved, suggesting the effectiveness of our proposed MDEO. As an exploratory work, MDEO has been successfully applied to networks of similar sizes. The knowledge transfer from a large network to a small network or vice versa, has not been studied yet. In future work, we will develop a more robust and adaptive framework that allows networks of various scales to be optimized simultaneously with improved effectiveness.

\bibliographystyle{IEEEtran}
\bibliography{zhao}

\vfill

\end{document}